\documentclass[letterpaper, 10 pt, conference]{ieeeconf}  % Comment this line out if you need a4paper

\IEEEoverridecommandlockouts                              % This command is only needed if
\usepackage{amsmath} % assumes amsmath package installed
\usepackage{amssymb}  % assumes amsmath package installed
\usepackage{graphicx}
\usepackage{siunitx} % to express SI units
\usepackage{balance}
\usepackage{url}
\usepackage{bm}
\usepackage[normalem]{ulem} % For strikethrough text
\usepackage[makeroom]{cancel} % For cancleing equations
\usepackage[dvipsnames]{xcolor}
\usepackage[font=small]{caption}
\usepackage{subcaption}
\usepackage{rotating}

\title{\LARGE \bf
Multi-task closed-loop inverse kinematics stability \\ through semidefinite programming}

\author{Josep Marti-Saumell, Angel Santamaria-Navarro, Carlos Ocampo-Martinez,~\IEEEmembership{Senior Member,~IEEE,}\\and Juan Andrade-Cetto% <-this % stops a space
\thanks{J. Marti-Saumell, C. Ocampo-Martinez and J. Andrade-Cetto are with the Institut de Rob\`otica i Inform\`atica Industrial, CSIC-UPC, Llorens Artigas 4-6, Barcelona 08028, Spain (e-mail: {jmarti, cocampo, cetto}@iri.upc.edu).}%
\thanks{A. Santamaria-Navarro is with NASA-JPL, California Institute of Technology, Pasadena, CA 91109 USA (e-mail: \mbox{angel.santamaria.navarro@jpl.nasa.gov}).}
\thanks{This work was partially supported by the EU H2020 project GAUSS (H2020-Galileo-2017-1-776293), project EB-SLAM (DPI2017-89564-P) and by the Spanish State Research Agency through the Mar\'ia de Maeztu Seal of Excellence to IRI (MDM-2016-0656).
Part of this research was carried out at the Jet Propulsion Laboratory, California Institute of Technology, under a contract with the National Aeronautics and Space Administration (NASA, USA).}
\thanks{This paper has a supplementary downloadable video, available at \mbox{http://ieeexplore.ieee.org}, showing experimental results of the presented ap
proach.}
}

\graphicspath{ {./figures/} }

\begin{document}

\maketitle
\thispagestyle{empty}
\pagestyle{empty}

%%%%%%% Mathematical Symbols ==>>

% Other commands
\newcommand{\inR}{\in\mathbb{R}}
\newcommand{\R}{\mathbb{R}}

\def\tr{^{\rm T}}
\def\trd{{}^{\rm T}}
\def\Atan{\mathrm{Atan2}}
\def\Acos{\mathrm{Acos}}
\def\sgn{\mathrm{sgn}}
\def\de{\mathrm{d}}
\def\diag{\mathrm{diag}}

\def\zero{\hbox{\bf 0}}
\def\Zero{{\mbox{\boldmath $O$}}}

\def\rchi{{\mbox{\raisebox{\depth}{$\chi$}}}}

% partial derivatives
\newcommand{\pdv}[2]{\mbox{$\frac{\partial #1}{\partial #2}$}}

% Dot product
\makeatletter
\newcommand*\dotp{\mathpalette\dotp@{.5}}
\newcommand*\dotp@[2]{\mathbin{\vcenter{\hbox{\scalebox{#2}{$\m@th#1\bullet$}}}}}
\makeatother

\def\bfa{\mathbf{a}}
\def\bfb{\mathbf{b}}
\def\bfc{\mathbf{c}}
\def\bfd{\mathbf{d}}
\def\bfe{\mathbf{e}}
\def\bff{\mathbf{f}}
\def\bfg{\mathbf{g}}
\def\bfh{\mathbf{h}}
\def\bfi{\mathbf{i}}
\def\bfj{\mathbf{j}}
\def\bfk{\mathbf{k}}
\def\bfl{\mathbf{l}}
\def\bfm{\mathbf{m}}
\def\bfn{\mathbf{n}}
\def\bfo{\mathbf{o}}
\def\bfp{\mathbf{p}}
\def\bfq{\mathbf{q}}
\def\bfr{\mathbf{r}}
\def\bfs{\mathbf{s}}
\def\bft{\mathbf{t}}
\def\bfu{\mathbf{u}}
\def\bfv{\mathbf{v}}
\def\bfw{\mathbf{w}}
\def\bfx{\mathbf{x}}
\def\bfy{\mathbf{y}}
\def\bfz{\mathbf{z}}
\def\bfA{\mathbf{A}}
\def\bfB{\mathbf{B}}
\def\bfC{\mathbf{C}}
\def\bfD{\mathbf{D}}
\def\bfE{\mathbf{E}}
\def\bfF{\mathbf{F}}
\def\bfG{\mathbf{G}}
\def\bfH{\mathbf{H}}
\def\bfI{\mathbf{I}}
\def\bfJ{\mathbf{J}}
\def\bfK{\mathbf{K}}
\def\bfL{\mathbf{L}}
\def\bfM{\mathbf{M}}
\def\bfN{\mathbf{N}}
\def\bfO{\mathbf{O}}
\def\bfP{\mathbf{P}}
\def\bfQ{\mathbf{Q}}
\def\bfR{\mathbf{R}}
\def\bfS{\mathbf{S}}
\def\bfT{\mathbf{T}}
\def\bfU{\mathbf{U}}
\def\bfV{\mathbf{V}}
\def\bfW{\mathbf{W}}
\def\bfX{\mathbf{X}}
\def\bfY{\mathbf{Y}}
\def\bfZ{\mathbf{Z}}

\def\bfGamma{\bm{\Gamma}}
\def\bfDelta{\bm{\Delta}}
\def\bfTheta{\bm{\Theta}}
\def\bfLambda{\bm{\Lambda}}
\def\bfXi{\bm{\Xi}}
\def\bfPi{\bm{\Pi}}
\def\bfSigma{\bm{\Sigma}}
\def\bfUpsilon{\bm{\Upsilon}}
\def\bfPhi{\bm{\Phi}}
\def\bfPsi{\bm{\Psi}}
\def\bfOmega{\bm{\Omega}}
\def\bfalpha{\bm{\alpha}}
\def\bfbeta{\bm{\beta}}
\def\bfgamma{\bm{\gamma}}
\def\bfdelta{\bm{\delta}}
\def\bfepsilon{\bm{\epsilon}}
\def\bfeta{\bm{\eta}}
\def\bftheta{\bm{\theta}}
\def\bfiota{\bm{\iota}}
\def\bfkappa{\bm{\kappa}}
\def\bflambda{\bm{\lambda}}
\def\bfmu{\bm{\mu}}
\def\bfnu{\bm{\nu}}
\def\bfxi{\bm{\xi}}
\def\bfpi{\bm{\pi}}
\def\bfrho{\bm{\rho}}
\def\bfsigma{\bm{\sigma}}
\def\bftau{\bm{\tau}}
\def\bfupsilon{\bm{\upsilon}}
\def\bfphi{\bm{\phi}}
\def\bfchi{\bm{\chi}}
\def\bfpsi{\bm{\psi}}
\def\bfomega{\bm{\omega}}
\def\bfvarepsilon{\bm{\varepsilon}}
\def\bfvartheta{\bm{\vartheta}}
\def\bfvarpi{\bm{\varpi}}
\def\bfvarrho{\bm{\varrho}}
\def\bfvarsigma{\bm{\varsigma}}
\def\bfvarphi{\bm{\varphi}}
\def\bfimath{\bm{\imath}}
\def\bfjmath{\bm{\jmath}}

\def\dbfa{\dot{\bfa}}
\def\dbfb{\dot{\bfb}}
\def\dbfc{\dot{\bfc}}
\def\dbfd{\dot{\bfd}}
\def\dbfe{\dot{\bfe}}
\def\dbff{\dot{\bff}}
\def\dbfg{\dot{\bfg}}
\def\dbfh{\dot{\bfh}}
\def\dbfi{\dot{\bfi}}
\def\dbfj{\dot{\bfj}}
\def\dbfk{\dot{\bfk}}
\def\dbfl{\dot{\bfl}}
\def\dbfm{\dot{\bfm}}
\def\dbfn{\dot{\bfn}}
\def\dbfo{\dot{\bfo}}
\def\dbfp{\dot{\bfp}}
\def\dbfq{\dot{\bfq}}
\def\dbfr{\dot{\bfr}}
\def\dbfs{\dot{\bfs}}
\def\dbft{\dot{\bft}}
\def\dbfu{\dot{\bfu}}
\def\dbfv{\dot{\bfv}}
\def\dbfw{\dot{\bfw}}
\def\dbfx{\dot{\bfx}}
\def\dbfy{\dot{\bfy}}
\def\dbfz{\dot{\bfz}}
\def\dbfA{\dot{\bfA}}
\def\dbfB{\dot{\bfB}}
\def\dbfC{\dot{\bfC}}
\def\dbfD{\dot{\bfD}}
\def\dbfE{\dot{\bfE}}
\def\dbfF{\dot{\bfF}}
\def\dbfG{\dot{\bfG}}
\def\dbfH{\dot{\bfH}}
\def\dbfI{\dot{\bfI}}
\def\dbfJ{\dot{\bfJ}}
\def\dbfK{\dot{\bfK}}
\def\dbfL{\dot{\bfL}}
\def\dbfM{\dot{\bfM}}
\def\dbfN{\dot{\bfN}}
\def\dbfO{\dot{\bfO}}
\def\dbfP{\dot{\bfP}}
\def\dbfQ{\dot{\bfQ}}
\def\dbfR{\dot{\bfR}}
\def\dbfS{\dot{\bfS}}
\def\dbfT{\dot{\bfT}}
\def\dbfU{\dot{\bfU}}
\def\dbfV{\dot{\bfV}}
\def\dbfW{\dot{\bfW}}
\def\dbfX{\dot{\bfX}}
\def\dbfY{\dot{\bfY}}
\def\dbfZ{\dot{\bfZ}}

\def\dbfGamma{\dot{\bfGamma}}
\def\dbfDelta{\dot{\bfDelta}}
\def\dbfTheta{\dot{\bfTheta}}
\def\dbfLambda{\dot{\bfLambda}}
\def\dbfXi{\dot{\bfXi}}
\def\dbfPi{\dot{\bfPi}}
\def\dbfSigma{\dot{\bfSigma}}
\def\dbfUpsilon{\dot{\bfUpsilon}}
\def\dbfPhi{\dot{\bfPhi}}
\def\dbfPsi{\dot{\bfPsi}}
\def\dbfOmega{\dot{\bfOmega}}
\def\dbfalpha{\dot{\bfalpha}}
\def\dbfbeta{\dot{\bfbeta}}
\def\dbfgamma{\dot{\bfgamma}}
\def\dbfdelta{\dot{\bfdelta}}
\def\dbfepsilon{\dot{\bfepsilon}}
\def\dbfzeta{\dot{\bfzeta}}
\def\dbfeta{\dot{\bfeta}}
\def\dbftheta{\dot{\bftheta}}
\def\dbfiota{\dot{\bfiota}}
\def\dbfkappa{\dot{\bfkappa}}
\def\dbflambda{\dot{\bflambda}}
\def\dbfmu{\dot{\bfmu}}
\def\dbfnu{\dot{\bfnu}}
\def\dbfxi{\dot{\bfxi}}
\def\dbfpi{\dot{\bfpi}}
\def\dbfrho{\dot{\bfrho}}
\def\dbfsigma{\dot{\bfsigma}}
\def\dbftau{\dot{\bftau}}
\def\dbfupsilon{\dot{\bfupsilon}}
\def\dbfphi{\dot{\bfphi}}
\def\dbfchi{\dot{\bfchi}}
\def\dbfpsi{\dot{\bfpsi}}
\def\dbfomega{\dot{\bfomega}}
\def\dbfvarepsilon{\dot{\bfvarepsilon}}
\def\dbfvartheta{\dot{\bfvartheta}}
\def\dbfvarpi{\dot{\bfvarpi}}
\def\dbfvarrho{\dot{\vbfarrho}}
\def\dbfvarsigma{\dot{\bfvarsigma}}
\def\dbfvarphi{\dot{\bfvarphi}}
\def\dbfimath{\dot{\bfimath}}
\def\dbfjmath{\dot{\bfjmath}}

\def\ddbfa{\ddot{\bfa}}
\def\ddbfb{\ddot{\bfb}}
\def\ddbfc{\ddot{\bfc}}
\def\ddbfd{\ddot{\bfd}}
\def\ddbfe{\ddot{\bfe}}
\def\ddbff{\ddot{\bff}}
\def\ddbfg{\ddot{\bfg}}
\def\ddbfh{\ddot{\bfh}}
\def\ddbfi{\ddot{\bfi}}
\def\ddbfj{\ddot{\bfj}}
\def\ddbfk{\ddot{\bfk}}
\def\ddbfl{\ddot{\bfl}}
\def\ddbfm{\ddot{\bfm}}
\def\ddbfn{\ddot{\bfn}}
\def\ddbfo{\ddot{\bfo}}
\def\ddbfp{\ddot{\bfp}}
\def\ddbfq{\ddot{\bfq}}
\def\ddbfr{\ddot{\bfr}}
\def\ddbfs{\ddot{\bfs}}
\def\ddbft{\ddot{\bft}}
\def\ddbfu{\ddot{\bfu}}
\def\ddbfv{\ddot{\bfv}}
\def\ddbfw{\ddot{\bfw}}
\def\ddbfx{\ddot{\bfx}}
\def\ddbfy{\ddot{\bfy}}
\def\ddbfz{\ddot{\bfz}}
\def\ddbfA{\ddot{\bfA}}
\def\ddbfB{\ddot{\bfB}}
\def\ddbfC{\ddot{\bfC}}
\def\ddbfD{\ddot{\bfD}}
\def\ddbfE{\ddot{\bfE}}
\def\ddbfF{\ddot{\bfF}}
\def\ddbfG{\ddot{\bfG}}
\def\ddbfH{\ddot{\bfH}}
\def\ddbfI{\ddot{\bfI}}
\def\ddbfJ{\ddot{\bfJ}}
\def\ddbfK{\ddot{\bfK}}
\def\ddbfL{\ddot{\bfL}}
\def\ddbfM{\ddot{\bfM}}
\def\ddbfN{\ddot{\bfN}}
\def\ddbfO{\ddot{\bfO}}
\def\ddbfP{\ddot{\bfP}}
\def\ddbfQ{\ddot{\bfQ}}
\def\ddbfR{\ddot{\bfR}}
\def\ddbfS{\ddot{\bfS}}
\def\ddbfT{\ddot{\bfT}}
\def\ddbfU{\ddot{\bfU}}
\def\ddbfV{\ddot{\bfV}}
\def\ddbfW{\ddot{\bfW}}
\def\ddbfX{\ddot{\bfX}}
\def\ddbfY{\ddot{\bfY}}
\def\ddbfZ{\ddot{\bfZ}}

\def\ddbfGamma{\ddot{\bfGamma}}
\def\ddbfDelta{\ddot{\bfDelta}}
\def\ddbfTheta{\ddot{\bfTheta}}
\def\ddbfLambda{\ddot{\bfLambda}}
\def\ddbfXi{\ddot{\bfXi}}
\def\ddbfPi{\ddot{\bfPi}}
\def\ddbfSigma{\ddot{\bfSigma}}
\def\ddbfUpsilon{\ddot{\bfUpsilon}}
\def\ddbfPhi{\ddot{\bfPhi}}
\def\ddbfPsi{\ddot{\bfPsi}}
\def\ddbfOmega{\ddot{\bfOmega}}
\def\ddbfalpha{\ddot{\bfalpha}}
\def\ddbfbeta{\ddot{\bfbeta}}
\def\ddbfgamma{\ddot{\bfgamma}}
\def\ddbfdelta{\ddot{\bfdelta}}
\def\ddbfepsilon{\ddot{\bfepsilon}}
\def\ddbfzeta{\ddot{\bfzeta}}
\def\ddbfeta{\ddot{\bfeta}}
\def\ddbftheta{\ddot{\bftheta}}
\def\ddbfiota{\ddot{\bfiota}}
\def\ddbfkappa{\ddot{\bfkappa}}
\def\ddbflambda{\ddot{\bflambda}}
\def\ddbfmu{\ddot{\bfmu}}
\def\ddbfnu{\ddot{\bfnu}}
\def\ddbfxi{\ddot{\bfxi}}
\def\ddbfpi{\ddot{\bfpi}}
\def\ddbfrho{\ddot{\bfrho}}
\def\ddbfsigma{\ddot{\bfsigma}}
\def\ddbftau{\ddot{\bftau}}
\def\ddbfupsilon{\ddot{\bfupsilon}}
\def\ddbfphi{\ddot{\bfphi}}
\def\ddbfchi{\ddot{\bfchi}}
\def\ddbfpsi{\ddot{\bfpsi}}
\def\ddbfomega{\ddot{\bfomega}}
\def\ddbfvarepsilon{\ddot{\bfvarepsilon}}
\def\ddbfvartheta{\ddot{\bfvartheta}}
\def\ddbfvarpi{\ddot{\bfvarpi}}
\def\ddbfvarrho{\ddot{\bfvarrho}}
\def\ddbfvarsigma{\ddot{\bfvarsigma}}
\def\ddbfvarphi{\ddot{\bfvarphi}}
\def\ddbfimath{\ddot{\bfimath}}
\def\ddbfjmath{\ddot{\bfjmath}}

% State vectors
\def\wpb{{\mbox{{${^w\bfp}_b$}}}}
\def\bpc{{\mbox{{${^b\bfp}_c$}}}}
\def\bpt{{\mbox{{${^b\bfp}_t$}}}}
\def\bpw{{\mbox{{${^b\bfp}_w$}}}}
\def\bpg{{\mbox{{${^b\bfp}_g$}}}}
\def\cpw{{\mbox{{${^c\bfp}_w$}}}}
\def\wphib{{\mbox{{${^w\bfphi}_b$}}}}
\def\cbfvartheta{{\mbox{{${^c\bfvartheta}$}}}}
\def\ibfvartheta{{\mbox{{${^i\bfvartheta}$}}}}
\def\tbfvartheta{{\mbox{{${^t\bfvartheta}$}}}}
\def\wbfvartheta{{\mbox{{${^w\bfvartheta}$}}}}

\def\dphi{{\mbox{{${\dot \phi}$}}}}
\def\dtheta{{\mbox{{${\dot \theta}$}}}}
\def\dpsi{{\mbox{{${\dot \psi}$}}}}

% Homogenous transforms and rotation matrices
\def\wTc{{\mbox{{${^w\bfT}\!_c$}}}}
\def\wTi{{\mbox{{${^w\bfT}\!_i$}}}}
\def\wTb{{\mbox{{${^w\bfT}\!_b$}}}}
\def\wTt{{\mbox{{${^w\bfT}\!_t$}}}}
\def\wTc{{\mbox{{${^w\bfT}\!_c$}}}}
\def\iTc{{\mbox{{${^i\bfT}\!_c$}}}}
\def\iTb{{\mbox{{${^i\bfT}\!_b$}}}}
\def\cTb{{\mbox{{${^c\bfT}\!_b$}}}}
\def\bTc{{\mbox{{${^b\bfT}\!_c$}}}}
\def\bTt{{\mbox{{${^b\bfT}\!_t$}}}}
\def\tTc{{\mbox{{${^t\bfT}\!_c$}}}}
\def\bRc{{\mbox{{${^b\bfR}_c$}}}}
\def\cRb{{\mbox{{${^c\!\bfR}_b$}}}}
\def\tRb{{\mbox{{${^t\!\bfR}_b$}}}}
\def\wRb{{\mbox{{${^w\!\bfR}_b$}}}}
\def\bRi{{\mbox{{${^b\!\bfR}_i$}}}}
\def\iRb{{\mbox{{${^i\!\bfR}_b$}}}}
\def\overcRb{{\mbox{{${^c\overline\bfR}_b$}}}}
\def\overtRb{{\mbox{{${^t\overline\bfR}_b$}}}}
\def\overwRb{{\mbox{{${^w\overline\bfR}_b$}}}}
\def\wRb{{\mbox{{${^w\!\bfR}_b$}}}}
\def\wRt{{\mbox{{${^w\!\bfR}_t$}}}}
\def\wRc{{\mbox{{${^w\!\bfR}_c$}}}}

% To be deleted
\def\bfvR{{\mbox{{$\bfv_{\!_R}$}}}}

% Jacobians
\def\obfJ{{\mbox{{$\overline{\bfJ}$}}}}
\def\JR{{\mbox{{$\bfJ_{\!\!_R}$}}}}
\def\JP{{\mbox{{$\bfJ_{\!\!_P}$}}}}
\def\JA{{\mbox{{$\bfJ_{\!\!_A}$}}}}

\newcommand{\sk}[1]{\lfloor#1\rfloor_{\!_\times}}

\def\mass{{\mbox{\rotatebox[origin=\depth]{180}{\raisebox{\depth}{$\omega$}}}}}

\newcommand{\proj}[1]{\bfP_{#1}}
\newcommand{\projt}[1]{\bfP^{'}_{#1}}

% \itemequation[label]{text before}{equation}
\makeatletter
\newcommand*{\itemequation}[3][]{%
  \item
  \begingroup
    \refstepcounter{equation}%
    \ifx\\#1\\%
    \else
      \label{#1}%
    \fi
    \sbox0{#2}%
    \sbox2{$\displaystyle#3\m@th$}%
    \sbox4{ \@eqnnum}%
    \dimen@=.5\dimexpr\linewidth-\wd2\relax
    % Warning for overlapping
    \let\CenterInSpace=N%
    \ifcase
        \ifdim\wd0>\dimen@
          \z@
        \else
          \ifdim\wd4>\dimen@
            \z@
          \else
            \@ne
          \fi
        \fi
      \let\CenterInSpace=Y%
    \fi
    \ifdim\dimexpr\wd0+\wd2+\wd4\relax>\linewidth
      \@latex@warning{Equation is too large}%
    \fi
    \noindent
    \rlap{\copy0}%
    \ifx\CenterInSpace Y%
      \rlap{\hbox to \linewidth{\kern\wd0\hss\copy2\hss\kern\wd4}}%
    \else
      \rlap{\hbox to \linewidth{\hfill\copy2\hfill}}%
    \fi
    \hbox to \linewidth{\hfill\copy4}%
    \hspace{0pt}% allow linebreak
  \endgroup
  \ignorespaces
}
\makeatother

\newcommand{\ie}{\emph{i.e.,~}}
\newcommand{\eg}{\emph{e.g.,~}}
\newcommand{\tdot}[1]{\dot{\tilde{#1}}}
\newcommand{\com}[1]{{\color{red}#1}}
\newcommand{\comb}[1]{{\color{blue}#1}}
\newcommand{\como}[1]{{\color{orange}#1}}
\newcommand{\figref}[1]{Fig. \ref{#1}}
\newcommand{\secref}[1]{Section \ref{#1}}

%%%%%%%%%%%%%%%%%%%%%%%%%%%%%%%%%%%%%%%%%%%%%%%%%%%%%%%%%%%%%%%%%%%%%%%%%%%%%%%
\begin{abstract}
Today's complex robotic designs comprise in some cases a large number of degrees of freedom, enabling for multi-objective task resolution (\eg humanoid robots or aerial manipulators).
This paper tackles the local stability problem of a hierarchical closed-loop inverse kinematics algorithm for such highly redundant robots.
We present a method to guarantee this system stability by performing an online tuning of the closed-loop control gains.
We define a semi-definite programming problem (SDP) with these gains as decision variables and a discrete-time Lyapunov stability condition as a linear matrix inequality, constraining the SDP optimization problem and guaranteeing the local stability of the prioritized tasks.
To the best of authors' knowledge, this work represents the first mathematical development of an SDP formulation that introduces these stability conditions for a multi-objective closed-loop inverse kinematic problem for highly redundant robots.
The validity of the proposed approach is demonstrated through simulation case studies, including didactic examples and a Matlab toolbox for the benefit of the community.
\end{abstract}

\section{Introduction}
Kinematically redundant robots, \eg ~\cite{kanoun_kinematic_2011, santamaria-navarro_task_2014}, have more degrees of freedom (DOFs) than those required to fulfill a particular task.
Finding the joint commands to fulfill a specific task, \ie solving the inverse kinematics (IK) problem, is usually done at a differential level and this causes a drift due to the eventual numerical integration.
To overcome this issue, one can resort to closed-loop inverse kinematic (CLIK) schemes~\cite{siciliano_closed-loop_1990,chiacchio_closed-loop_1991}, which consist in finding proper joint values such that the task errors are driven towards zero.
Although most IK algorithms can be easily modified to become CLIK methods, there exist some implications when solving for several tasks.

A common approach to solve multiple tasks simultaneously for a redundant robot is to introduce task priorities while combining them in a single control law.
Hence, if the robot cannot fulfill all tasks, it can prioritize the solution of those placed at the top of the hierarchy.
The technique presented in~\cite{nakamura_task-priority_1987}, satisfies lower priority tasks only in the null space of the higher priority ones.
A similar approach is taken in~\cite{siciliano_general_1991}, this time using task-augmented Jacobians.
By using these approaches, when two tasks are not independent (\ie when they share their corresponding null-space), the algorithm suffers from algorithmic singularities, leading to unstable joint velocities~\cite{chiaverini_singularity-robust_1997}.
In~\cite{chiaverini_singularity-robust_1997}, an algorithmic singularity robust method is proposed to solve for two tasks separately using a classical least-squares method. 
The conflict between tasks is filtered out by projecting the second task solution into the null-space of the first one. 
This approach is analyzed and extended for several tasks in~\cite{baerlocher_task-priority_1998}, where the projection is done in the augmented null space.
Although this null-space technique has been used in recent works like~\cite{santamaria-navarro_task_2014,baizid_behavioral_2017} or~\cite{santamaria-navarro_uncalibrated_2017}, they usually lack of a rigorous formulation development.
For instance, all these methods are usually developed in continuous time while afterward they are implemented in discrete time, hence bypassing the influence of the sampling time into the overall system's performance.
Besides, the overall control law stability has not been analyzed and just analysis for individual tasks is presented, without a mention of the instability behaviors that can arise while combining them.

The stability of CLIK problems, might be affected by tasks dependencies, \ie the completion of a particular task might prohibit the fulfillment of another eventual task.
However, there exists the case where all task errors behave as desired if the proper closed-loop control gains are selected.
In that sense, \cite{antonelli_stability_2009} presents the dependency between tasks and gains together with a way to measure it.
Besides, it studies the local stability of a task-priority CLIK problem taking advantage of the Lyapunov theory, which allows finding conservative upper and lower bounds for the task gains.
Unfortunately, the analytical developments presented in \cite{antonelli_stability_2009} are focused on three tasks and become intractable when extended to $N$ tasks.
A similar problem appears in~\cite{falco_stability_2011} where stability conditions are only provided for a single task.
An extension of~\cite{antonelli_stability_2009} is~\cite{moe_stability_2015}, where the local stability of systems dealing with inequality tasks is analyzed. 
However, they assume a manually chosen set of gains and there is no insight provided on how to compute them.
The gain-tuning solution is one of the main novelties presented hereafter in this paper.

Most of the existing methods that analyze the use of multiple tasks in a hierarchy are presented in continuous-time formulations~\cite{antonelli_stability_2009, moe_stability_2015}.
Discrete-time CLIK schemes add another factor to consider when proving the stability of the system: the sampling time selection.
This effect is studied in~\cite{das_inverse_1988} in the sense of Lyapunov theory, and in~\cite{falco_stability_2011} without resorting to such theory.
However, in both cases, the analysis refers to single tasks.

The novel contribution of this paper is a method to find optimal task feedback gains for the discrete hierarchical CLIK regulation problem, guaranteeing local stability of all tasks in the hierarchy.
We choose a singularity free approach~\cite{baerlocher_task-priority_1998}, describe it as a discrete-time CLIK system and, taking advantage of an SDP problem definition, we find the optimal gains which guarantee system stability.
% Notice that the convergence of the SDP problem is guaranteed as it is convex by definition.
With the SDP approach defined hereafter, local stability can be guaranteed by adding a constraint to the optimization problem.
This constraint is formulated in the sense of Lyapunov and as a linear matrix inequality (LMI), where the desired gains are the optimization variables.
Apart from guaranteeing local stability, this constraint allows us to modify the error dynamics, \ie to get faster or slower error convergence towards zero.
Besides, in the SDP we can account for the sampling time and also add further conditions, such as to limit the joint velocities.
% To the best of our knowledge, this is the first work that uses an SDP formulation to introduce local stability conditions in the solution of a hierarchical CLIK problem for highly redundant robots.
We stress that this work addresses the \emph{local} stability of the system as the global stability for multiple task hierarchical resolution is still an open question for discrete-time systems and we consider it out of the paper scope.
For the sake of simplicity, in the rest of the paper we refer as stability to this local stability.

% The remainder of this article is structured as follows.
% In Section \ref{sec:background}, we describe the required background, including the hierarchical IK and CLIK formulations.
% The system stability developments are presented in Section \ref{sec:stability}.
% Section \ref{sec:sdp} develops the formulation of the SDP problem to perform an online tuning of the task gains while guaranteeing closed-loop stability, together with additional constraints.
% The validation of the proposed method is illustrated in Section \ref{sec:validation}.
% Finally, discussion and conclusions are drawn in Section \ref{sec:conclusions}.

\section{Background}
\label{sec:background}
This section introduces the required background formulation related to IK and CLIK algorithms.
At the end we state the stability conditions that we must ensure with the proposed method.

\subsection{Hierarchical inverse kinematics} \label{subsec:inverse_kinematics}

Our formulation has drawn inspiration from the algorithmic singularity-free IK presented in~\cite{baerlocher_task-priority_1998}.
Let us define an $i$-th task $\bfsigma_i(t) \inR^{n_i}$, as a function of the robot joints,
\begin{equation}
\label{eq:task_def}
\bfsigma_i(t) = \bff_i(\bfq(t)) \,,
\end{equation}
being $\bfq \inR^\nu$ the joint values, \ie the robot configuration.
Solving the IK problem consists of solving the inverse of \eqref{eq:task_def}.
As said before, this is done at a differential level. Therefore,
\begin{equation}
\label{eq:ik_minnorm_solution}
\dot{\bfq} = \bfJ^\dagger_i\dot{\bfsigma}_i \,,
\end{equation}
where $\bfJ_i\inR^{n_i\times \nu}$ is the Jacobian matrix of the task in \eqref{eq:task_def} and ${\bfJ^\dagger_i = \bfJ^\top_i(\bfJ_i\bfJ^\top_i)^{-1} \inR^{\nu\times n_i}}$ is its Moore-Penrose pseudo-inverse.
Here, we assume to be working in a region free from kinematic singularities, 
hence $\bfJ_i$ will be full rank and \eqref{eq:ik_minnorm_solution} does not make use of the damped pseudo-inverse as in~\cite{baerlocher_task-priority_1998}.

In order to accomplish a secondary task simultaneously while imposing a hierarchy, one can take advantage of the motions residing in the null space of the primary task.
A first work presenting this technique is~\cite{nakamura_task-priority_1987} where joint velocities for the secondary task are computed so as not to modify the primary task.
However, as analyzed in~\cite{chiaverini_singularity-robust_1997}, this method suffers from algorithmic singularities and proposes a solution in which tasks at two different hierarchy levels are solved separately.
Then, the low priority task solution is projected onto the null space of the task higher in the hierarchy.
This technique is analyzed and generalized to more than two priority levels in~\cite{baerlocher_task-priority_1998}.
Thus, in the case of having $h$ hierarchy levels, the solution to the IK problem results in
\begin{equation}
\label{eq:ik_chiaverini}
\dot{\bfq} = \bfJ_1^\dagger\dot{\bfsigma}_1 + \overline{\bfN}_1\bfJ_2^\dagger\dot{\bfsigma}_2 + \dots + \overline{\bfN}_{h-1}\bfJ_{h}^\dagger\dot{\bfsigma}_{h}\,,
\vspace{-0.5em}
\end{equation}
where $\overline{\bfN}_{h-1} = \bfI_n - \bfJ_{1\dots h-1}^\dagger \bfJ_{1\dots h-1}$ is the null-space projector of the augmented Jacobian matrix
\begin{equation}
  \bfJ_{1\dots h-1} = [\bfJ_1^\top \quad \bfJ_2^\top \dots  \bfJ_{h-1}^\top]^\top \,,
  \vspace{-0.5em}
\end{equation}
with ${\bfJ_{1\dots h-1}\inR^{(n_1+\dots+n_{h-1})\times \nu}}$.

\subsection{Closed-loop inverse kinematics}
The aforementioned IK solution has to be computed in the discrete-time domain.
Thus, given a trajectory in the task-space we obtain its analogous in the joint space by numerical integration, \eg by using a first order Euler integration
\begin{equation}
 \bfq^{(k+1)} = \bfq^{(k)} + \dbfq^{(k)}\Delta t \,,
 \label{eq:joint_euler_integration}
\end{equation}
where $\bfq^{(k)} = \bfq(t_k)$ and $t_k$ is the time at integration step $k$.
We use $\dot{\Box}^{(k)}$ to express the velocity of a variable evaluated at time $t_k$.
This IK discrete implementation entails a drifting problem provoked by numerical integration.
To overcome it, we can formulate a closed-loop version of the IK problem (CLIK). The continuous-time version of the closed-loop solution defines the task error and assigns a dynamics to it as
\begin{subequations}
\begin{equation}
  \label{eq:clik_error}
%   \vspace{-0.5em}
  \tilde{\bfsigma}_i = \bfsigma_i^\ast - \bfsigma_i \,,
  \vspace{-0.5em}
\end{equation}
\begin{equation}
  \label{eq:error_dynamics}
  \tdot{\bfsigma}_i = -\bfLambda_i\tilde{\bfsigma}_i  \,,
  \vspace{-0.5em}
\end{equation}
\end{subequations}
where $\bfsigma_i^\ast \inR^{n_i}$ is the desired task value.
In order to decrease the error towards zero, $\bfLambda_i \inR^{n_i\times n_i}$ is a positive-definite diagonal matrix of suitable gains.
Differentiating~\eqref{eq:clik_error} with respect to time, combining it with~\eqref{eq:error_dynamics} and isolating $\dot{\bfsigma}_i$,
we can directly substitute $\dot{\bfsigma}_i = \dbfsigma_i^\ast + \bfLambda_i\tilde{\bfsigma}_i$ into \eqref{eq:ik_chiaverini}.
Hence, the analogous equation for a CLIK problem with $h$ priority levels becomes
\begin{equation}
    \vspace{-0.5em}
  \label{eq:clik_chiaverini}
  \dot{\bfq} = \bfJ_1^\dagger\bfLambda_1\tilde{\bfsigma}_1 + \overline{\bfN}_1\bfJ_2^\dagger\bfLambda_2\tilde{\bfsigma}_2
  + ... + \overline{\bfN}_{h-1}\bfJ_{h}^\dagger\bfLambda_h\tilde{\bfsigma}_h \,.
%   \vspace{-0.5em}
\end{equation}
Notice how \eqref{eq:clik_chiaverini} includes different desired values and gain matrices for each task.
Finally, when considering the discrete-time system, we can obtain $\dbfq^{(k)}$ from~\eqref{eq:joint_euler_integration} by evaluating the multiple Jacobian matrices and task errors in~\eqref{eq:clik_chiaverini} at time $t_k$.

This discrete-time CLIK formulation is considered in sections \ref{sec:stability} and \ref{sec:sdp} to state an SDP problem 
that, when solved, outputs the gains that render all tasks stable.
% Although all equations stated hereafter consider the discrete-time domain (\ie matrices and vectors must be evaluated at the corresponding $t_k$), for the sake of conciseness, the super-index $(k)$ indicating the evaluation time will be only written when confusion may occur. 
For the sake of conciseness, the super-index $(k)$ indicating the evaluation time will be only written when confusion may occur. 

%%%%%%%%%%%%%%%%%%%%%%%%%%%%%%%%%%%%%%%%%%%%%%%%%%%%%%%%%%%%%%%%%%%%%%%%%%%%%%%%
%%%%%%%%%%%%%%%%%%%%%%%%%%%%%%%%%%%%%%%%%%%%%%%%%%%%%%%%%%%%%%%%%%%%%%%%%%%%%%%%
%%%%%%%%%%%%%%%%%%%%%%%%%%%%%%%%%%%%%%%%%%%%%%%%%%%%%%%%%%%%%%%%%%%%%%%%%%%%%%%%

\section{Stability analysis}
\label{sec:stability}
By definition, a single task $i$ is stable if its error decreases asymptotically to zero.
However, when a hierarchy is applied to solve several tasks simultaneously for a redundant robot, 
the interaction between tasks affects the overall control law stability.
As stated in~\cite{antonelli_stability_2009}, the stability of the closed-loop controlled scheme can be guaranteed if we assume independent tasks, as done in this work, and with proper tuning of the task gains $\bfLambda$.
In the following, we present one of the key novelties of this paper: the condition to guarantee the overall stability considering a discrete-time system, which depends on these task gains.
As the analytical computation of the gains for $N$ tasks is unfeasible \cite{antonelli_stability_2009}, the condition presented here will be later on introduced as a constraint in an SDP optimization procedure.

In order to analyze the stability of the whole system, let us consider an augmented vector containing all task errors:
\begin{equation}
  \label{eq:error_augmented}
  \tilde{\bfsigma}^\top = [\tilde{\bfsigma}_{1}^\top  \dots \tilde{\bfsigma}_{h}^\top] \,,
\end{equation}
being $\tilde{\bfsigma} \inR^{n}$ the augmented error, with $n = n_1 + \dots + n_h$ at time $t_k$.
We can assess the stability of a system by resorting to the Lyapunov theory for discrete systems.
Given a Lyapunov candidate function \mbox{$V(\tilde{\bfsigma}^{(k)}) = V^{(k)} > 0\,$,} $\text{ } \forall \tilde{\bfsigma}^{(k)} \neq \bf0$, the error will decrease towards zero if \mbox{$V^{(k+1)} - V^{(k)} < 0$} holds, \ie if the Lyapunov candidate decreases with time~\cite{boyd_linear_1994}.
We choose it to be 
\begin{equation}
  V(\tilde{\bfsigma}) = \frac{1}{2} \tilde{\bfsigma}^\top\tilde{\bfsigma} \,.
  \label{eq:lyap_unov_candidate}
\end{equation}
Hence, to guarantee the stability of the system we must ensure that
\begin{equation}
  \cfrac{1}{2}\tilde{\bfsigma}^{(k+1)\top}\tilde{\bfsigma}^{(k+1)} - \cfrac{1}{2}\tilde{\bfsigma}^{(k)\top}\tilde{\bfsigma}^{(k)}< 0 \,,
  \label{eq:lyapunov_error}
\end{equation}
where the error $\tilde{\bfsigma}^{(k+1)}$ can be approximated by a Taylor series expansion of $\tilde{\bfsigma}(t)$ around $t_k$ and evaluating it at $t_{k+1}$ up to the first term (first order Euler integration).
Thus,
\begin{equation}
  \tilde{\bfsigma}^{(k+1)} \approx \tilde{\bfsigma}^{(k)} + \tdot{\bfsigma}^{(k)}\Delta t \,.
  \label{eq:taylor_error}
\end{equation}
As shown in~\cite{das_inverse_1988}, this approximation is valid as long as the higher-order terms remain small. According to this work, the higher-order terms can be neglected if $||\dbfq||\Delta t$ is below a certain bound.
To fulfill this statement, we add another constraint to our SDP problem, as explained in the following section.
Notice that guaranteeing \eqref{eq:lyapunov_error} implies local stability depending on the tasks gains, the sampling time and the initial value of the error.
In this paper, we propose a solution considering the first two factors and assume, as commonly done in the literature, an initial value of the error that can keep the problem feasible.
In \cite{falco_stability_2011}, it is proposed a method to estimate the region of attraction for a single task, depending on the task gain selected, the sampling time and several parameters related to the derivatives of $\bff(\bfq(t))$.
However, an estimation of the region of attraction for several tasks still remains as an open problem, hence we limit the scope of the paper to guarantee local stability.

Now we can substitute~\eqref{eq:taylor_error} in~\eqref{eq:lyapunov_error} obtaining
\begin{equation} \label{eq:lyapunov_approx}
\begin{split}
  V^{(k+1)} - V^{(k)} & \approx
  \cfrac{1}{2}
  (
    \tdot{\bfsigma}^{(k)\top} \tilde{\bfsigma}^{(k)}\Delta t + \tilde{\bfsigma}^{(k)\top}\tdot{\bfsigma}^{(k)}\Delta t + \\
    &+ \tdot{\bfsigma}^{(k)\top}\tdot{\bfsigma}^{(k)}\Delta t^2
  ) \,,
  \end{split}
\end{equation}
which no longer depends on $\tilde{\bfsigma}^{(k+1)}$ and therefore, from now on, the super-index $k$ will be omitted.
In order to keep developing~\eqref{eq:lyapunov_approx}, we use the following expression for the error velocity evaluated at time $t_k$,
\begin{equation}
  \tdot{\bfsigma} = [\tdot{\bfsigma}_{1}^\top~\hdots~\tdot{\bfsigma}_{h}^\top
  ]^\top
  =
  - [\bfJ_{1}^\top~ \hdots~ \bfJ_{h}^\top]^\top
  \dot{\bfq}
  \,.
  \label{eq:error_velocity}
\end{equation}
Notice how this expression can be obtained by differentiating \eqref{eq:clik_error} and combine it with $\dot{\bfsigma}_i = \bfJ_i \dot{\bfq}$.
This can be further expanded by using the CLIK solution in~\eqref{eq:clik_chiaverini} resulting in the following linear mapping
\begin{align}
  \tdot{\bfsigma}
  =
  \begin{bmatrix}
  -\bfJ_1\bfJ_1^\dagger\bfLambda_1 & \cdots & -\bfJ_1\overline{\bfN}_{h-1}\bfJ_h^\dagger\bfLambda_h \\
  \vdots & \ddots & \vdots \\
  -\bfJ_h\bfJ_1^\dagger\bfLambda_1 & \cdots & -\bfJ_h\overline{\bfN}_{h-1}\bfJ_h^\dagger\bfLambda_h \\
  \end{bmatrix}
  \tilde{\bfsigma} = \bfA \tilde{\bfsigma}\,.
  \label{eq:error_velocity_error}
\end{align}
Now, by substituting~\eqref{eq:error_velocity_error} into~\eqref{eq:lyapunov_approx}, we obtain the stability condition of the stack of tasks, which will be guaranteed if we can assure the positive definiteness of the expression
\begin{equation}
  - \bfA^\top\Delta t - \bfA\Delta t - \bfA^\top\bfA\Delta t^2 \succ 0 \,,
  \label{eq:stability_condition}
\end{equation}
leading the candidate Lyapunov function to become an actual Lyapunov function.
The symbol "$\succ$" stands for positive definite matrix.

System stability is guaranteed if we can find proper gain values $\bfLambda_i$ so that condition~\eqref{eq:stability_condition} holds at every time step.
In this paper, we choose these gains as the decision variables in the SDP optimization problem and, by performing an online gain tuning, we account for the error dynamic change at every specific robot configuration.

\section{SDP-based gain scheduling}
\label{sec:sdp}

% \comb{AS: subscript notation: }

% \comb{

% 	$i$: idx task with $i \in [1,k]$

% 	$l$: idx $\lambda$ with $l \in [1,n]$

% 	$j$: idx LMIs with $j \in [1,m]$
% }
We want to formulate an SDP problem with the stability condition~\eqref{eq:stability_condition} as an LMI constraint.
So, this SDP will be used to find the optimal closed-loop control gains that guarantee the stability of the error in~\eqref{eq:error_augmented}.
An SDP problem is a convex optimization problem whose feasible set is a cone formed by positive semidefinite symmetric matrices~\cite{boyd_convex_2004,wolkowicz_handbook_2000}.
Making use of LMIs, this kind of problems allows us to impose constraints on the definiteness of matrices.
They have the following form:
% \begin{equation}
% \label{eq:SDP_standard_form}
% \begin{array}{rcl}
% \displaystyle \min_{\bfx} 	& \hspace{-1.5em} \bfc^\top \bfx  \,,
% \textrm{s.t.}             	& \hspace{0.5em} \bfF(\bfx) \succeq 0 \,,
% \end{array}
% \end{equation}
\begin{equation}
\label{eq:SDP_standard_form}
\begin{array}{rclll}
\displaystyle \min_{\bfx} 	& \bfc^\top \bfx  & \textrm{s.t.} & \bfF(\bfx) \succeq 0 \,,
\end{array}
\end{equation}
where $\bfx = [x_1, \dots, x_r]^\top \inR^r$ is the vector of decision variables, $\bfc\inR^r$ is a vector of coefficients and \mbox{$\bfF(\bfx)\inR^{s\times s}$} is a positive semi-definite LMI (noted with "$\succeq$").
The dimensions $r$ and $s$ will be defined hereafter.

Our goals in specifying the 
% \sout{constraint} 
elements of~\eqref{eq:SDP_standard_form} 
% \sout{(LMIs)} 
are to impose closed-loop stability and limit the resulting joint velocities, so the approximation in \eqref{eq:taylor_error} holds, while trying to impose a convergence speed.
% \sout{and setting the gains to specific values (if possible).}
Each of these constraints will be defined as single LMIs $\bfF_j(\bfx)$, described in the following subsections.
Afterwards, all these single LMIs will be formulated as an LMI of the form $\bfF(\bfx)$ by placing them into a block diagonal matrix
\begin{equation}
\label{eq:fx_blockDiag}
\bfF(\bfx) = \mathsf{diag}(\bfF_1(\bfx), \dots, \bfF_m(\bfx)) \succeq 0\,,
\end{equation}
with $m$ the number of single LMIs.

The optimized outputs of the SDP problem are the $\bfLambda_i$ gain matrices, which have the form
\begin{equation}
  \label{eq:lambda_i}
  \bfLambda_i = \mathsf{diag}(\lambda_{i,1}, \ldots, \lambda_{i,n_i}) \,, \text{ for } i=1,\dots,h \,,
\end{equation}
thus the ${\bflambda_i = [\lambda_{i,1},...,\lambda_{i,n_i}]^\top}$ vectors will be part of the decision variable $\bfx$ in \eqref{eq:SDP_standard_form}.
Therefore, each single LMI $\bfF_j(\bfx)$ will have the form
\begin{multline}
% \begin{split}
\label{eq:single_LMI}
\bfF_j(\bfx) = \bfF_{j,0} + \bfF_{j,1}\lambda_{i,1} + \cdots + \bfF_{j,n_i}\lambda_{i,n_i} \succeq 0 \,,
% \end{split}
\end{multline}
% In the following subsections, we describe in detail the form of each of these single LMIs $\bfF_j(\bfx)$ that form the SDP constraint $\bfF(\bfx)$ in \eqref{eq:fx_blockDiag}.
for $i=1,\dots,h $. 
For the sake of simplicity, in the following we join all task gains in a single vector of $n$ elements, \ie ${\bflambda = [\lambda_{1,1},...,\lambda_{1,n_1},...,\lambda_{h,1},...\lambda_{h,n_h}]^\top \inR^n}$ with ${n = \sum_{s=1}^h n_s}$, being the sum of task dimensions.
% Besides, notice that online gain tuning consists of solving the SDP problem at each iteration, obtaining different $\bflambda$ at each $t_k$.

\subsubsection{$\bfF_1$: Stability} \label{sec:F1}
In order to express~\eqref{eq:stability_condition} as an LMI, we require some mathematical manipulations.
On the one hand, notice that~\eqref{eq:stability_condition} imposes the strict definiteness ($\succ$) in contrast to the semidefiniteness ($\succeq$) required by LMIs in~\eqref{eq:SDP_standard_form}.
We can impose strict positive definiteness by considering a scalar factor $\beta>0$ such that, when multiplied by the identity of suitable dimensions, expression ~\eqref{eq:stability_condition} becomes
% To accomplish with this restriction, we can add a scalar factor $\alpha>0$ that when multiplied by the identity of suitable dimensions, will impose the strict positive definitness on~\eqref{eq:stability_condition}, \ie
% \begin{equation}
% - \bfA^\top \Delta t - \bfA \Delta t - \bfA^\top\bfA \Delta t^2 \succeq \alpha \bfI \,.
% \label{eq:stability_condition_mod}
% \end{equation}
% Furthermore, $\alpha$ can be used to modify the error dynamics, so \comb{the higher the value of $\alpha$ is, the faster the errors' convergence will be} \sout{depending on its value the speed of error convergence will vary}.
% Notice that, if we consider the scalar $\beta \triangleq \frac{\alpha}{\Delta t}$,~\eqref{eq:stability_condition_mod} can be further simplified, obtaining
\begin{equation}
- \bfA^\top - \bfA - \bfA^\top \bfA \Delta t \succeq \beta \bfI \,.
\label{eq:stability_condition_mod_2}
\end{equation}
Note also how $\beta$ can be used to modify the error dynamics, so the higher the value of $\beta$, the faster the errors will convergence.
However, setting $\beta$ too high can jeoparadize the SDP feasibility as it is contradictory with limiting the joints velocities.
To overcome this issue, we set a soft equality constraint on $\beta$ in the SDP problem (see \secref{sec:beta_soft_constraint} for implementation details).

Notice how \eqref{eq:stability_condition_mod_2} depends quadratically on the task gains.
To express it as an LMI (linear dependence on gains), we take advantage of the Schur complement for symmetric matrices.
% First, let us define a symmetric matrix of suitable dimensions as
% \begin{equation}
%   \bfM =
%   \begin{bmatrix}
%     \bfB & \bfC \\
%     \bfC^\top & \bfD
%   \end{bmatrix} \,.
% \end{equation}
% The Schur complement condition for positive definiteness states that, considering $\bfD$ is positive definite, $\bfM$ is positive semi-definite \emph{if and only if} its Schur complement $\bfM|\bfD$ is positive semi-definite.
% Hence,
% \begin{equation}
%   \text{If } \bfD \succ 0, \text{ then } \bfM \succeq 0 \Leftrightarrow \bfM|\bfD = \bfB - \bfC\bfD^{-1}\bfC^\top \succeq 0\,.
% \end{equation}
Then, doing the proper assignments, the expression~\eqref{eq:stability_condition_mod_2} becomes the Schur complement of the matrix
\begin{equation}
    \bfM =
    \begin{bmatrix}
    -(\bfA^\top + \bfA)- \beta \bfI & \bfA^\top \Delta t^{1/2} \\
    \bfA\Delta t^{1/2} & \bfI
    \end{bmatrix} \succeq 0 \,,
\end{equation}
which finally, depends linearly on the gains $\bflambda_i$.

The details on how to express $\bfM$ in terms of $\bfLambda$, $\bfM(\bfLambda)$, are presented in the following.
This procedure is only detailed for the block ($\bfA^\top + \bfA$).
The developments for the rest of the blocks are straight-forward and are here omitted for the sake of brevity.
Basically, we express the matrix $\bfA$ in terms of the different gain matrices $\bfLambda_i$ to obtain the following expression
\begin{multline}
\label{eq:A_At_lambda}
\bfA^\top(\bfLambda) + \bfA(\bfLambda) =
\\
\begin{bmatrix}
\bfA_{1,1} \bfLambda_1 + \bfLambda_1\bfA_{1,1}^\top & \cdots &  \bfA_{1,h}\bfLambda_h + \bfLambda_1 \bfA_{h,1}^\top  \\
\vdots & \ddots & \vdots \\
\bfA_{h,1}\bfLambda_1 + \bfLambda_h \bfA_{1,h}^\top  & \cdots & \bfA_{h,h}\bfLambda_h + \bfLambda_h \bfA_{h,h}^\top
\end{bmatrix} \,.
\end{multline}
being ${\bfA_{i,\rho} = -\bfJ_i\overline{\bfN}_{\rho-1}\bfJ_\rho^\dagger}$ for $i,\rho = 1,\dots,h$ from~\eqref{eq:error_velocity_error}, with $\overline{\bfN}_0 = \bfI$.

% \begin{equation}
%   \label{eq:A_ofLambda}
%   \bfA(\bfLambda) =
%   \begin{bmatrix}
%   \bfA_{1,1} & \cdots & \bfA_{1,h} \\
%   \vdots & \ddots & \vdots \\
%   \bfA_{h,1} & \cdots & \bfA_{h,h}
%   \end{bmatrix}
%   \begin{bmatrix}
%   \bfLambda_1 & & \bf0  \\
%   & \ddots &  \\
%   \bf0 &  & \bfLambda_h
%   \end{bmatrix} \,,
% \end{equation}

% Then, using~\eqref{eq:A_ofLambda} we obtain the symmetric matrix

The stability LMI requires $\bfM(\bfLambda)$ to be positive definite, \ie ${\bfF_1(\bflambda) = \bfM(\bfLambda)}$.
To express $\bfM$ in terms of the corresponding vector $\bflambda$, which is part of the decision variables, the matrices $\bfF_{1,l}$ with $l\in [1,n]$ in \eqref{eq:single_LMI} must be the elements of $\bfM$ that are multiplied by the corresponding gains in $\bflambda$.
Notice that this LMI also takes into account the sampling time $\Delta t$, whose effect is shown in \secref{sec:validation}.

\subsubsection{$\bfF_2$: Joint velocity limits}
In order to make the approximation in~\eqref{eq:taylor_error} hold, we must bound the joint velocity so second-order terms of the Taylor expansion do not become significant.
With that aim, we resort to \eqref{eq:clik_chiaverini} to express every component of $\dot{\bfq} = [\dot{q}_1, \dots, \dot{q}_n]^\top$ as a linear combination of the gains, hence we can transform \eqref{eq:clik_chiaverini} into
\begin{equation}
  \label{eq:jl_1}
  \dot{\bfq} = \bfJ_1^\dagger \bfSigma_1 \bflambda_{1} + \dots +
  \overline{\bfN}_{h-1}\bfJ^\dagger_h \bfSigma_h\bflambda_h \,,
\end{equation}
where we express every $\bfLambda_i$ in its vector form and the error vector of every task is replaced by a diagonal matrix, \ie $\bfSigma_h = \mathsf{diag}(\tdot{\bfsigma}_h)$.
This expression can be rearranged as
\begin{equation}
	\dot{\bfq} = \bfS\bflambda \,,
\end{equation}
with $\bfS \triangleq
  \begin{bmatrix}
  \bfJ_1^\dagger\bfSigma_1 | \cdots | \overline{\bfN}_{h-1}\bfJ^\dagger_h \bfSigma_h
  \end{bmatrix}$.
% \begin{equation}
%   \label{eq:jl_Smatrix}
%   \bfS \triangleq
%   \begin{bmatrix}
%   \bfJ_1^\dagger\bfSigma_1 | \cdots | \overline{\bfN}_{h-1}\bfJ^\dagger_h \bfSigma_h
%   \end{bmatrix}\,.
% \end{equation}
Then, we can add upper bounds to the joints velocities considering
\begin{equation}
  \label{eq:jl_ub}
  \overline{\dot{\bfq}} - \bfS \bflambda \geq 0 \,,
\end{equation}
where $\overline{\dot{\bfq}}$ is the vector containing the upper bounds, and the symbol $\geq$ stands for the element-wise operand $\geq$.
Finally, we can convert \eqref{eq:jl_ub} into an LMI of the form $\bfF \succeq 0$ by specifying
\begin{subequations}
	\begin{align}
		\bfF_{2,0} &= \mathsf{diag}(\overline{\dot{\bfq}}) \,, \\
		\bfF_{2,j} &= -\mathsf{diag}(\bfs_j) \text{ for } j=1,\dots,n \,,
	\end{align}
\end{subequations}
being $\bfs_j$ the $j$-th column of $\bfS$.
Notice that ${\bfF_{2,j} \inR^{\nu\times\nu}}$.

The addition of lower bounds is done analogously to the upper bounds procedure and has been omitted here for the sake of brevity.

\subsubsection{$\bfF_3$: Soft constraint on $\beta$} \label{sec:beta_soft_constraint}
The variable $\beta$ will direct the velocity of the error convergence in \eqref{eq:stability_condition_mod_2}.
Although high values of $\beta$ may lead to fast convergence, this behavior might go against limiting the joints' velocities with the constraint described above, thus it difficults the convergence of the SDP problem.
To avoid this, we can define a soft constraint on $\beta$ with an initial desired value $\tilde{\beta}$, such that
% As it has been briefly noted before, setting a high value for $\beta$ may be contradictory to limit the joints' velocity, as it can difficult the convergence of the SDP problem.
% Thus, $\beta$ will not be set as a fixed parameter at the begining of the algorithm.
% Instead, we will set a desired beta $\tilde{\beta}$.}
% If possible, the SDP problem will output the gains that accomplish with the speed convergence of $\tilde{\beta}$.
% If this represents a conflict with the stability and joint velocity requirements, the constraint on $\beta$ will be relaxed.
% So far, we have only defined LMIs that need to be satisfied (constraints), which form a subspace of feasible solutions.
% However, we have not specified a criterion to choose one specific solution contained in this subspace.
% With that aim, we propose that, 
% \sout{given a vector of desired gains $\tilde{\bflambda}\inR^n$, we try to find the closest set inside the feasible subspace defined by $\bfF_1(\bflambda)$ and $\bfF_2(\bflambda)$} 
% The optimization problem considering  $\tilde{\beta}$ is written as:
% \begin{equation}
% \xcancel{\begin{array}{rcl}
% \displaystyle \min   & \hspace{-0.5em} ||\bflambda -\tilde{\bflambda}||^2 \,,\\
% \textrm{s.t.}        & \hspace{0em} \bfF_1,\bfF_2 \succeq 0 \,.
% \end{array}}
% \end{equation}
% \vspace{-0.4em}
% \begin{equation}
% % \vspace{-0.4em}
%   \label{eq:sdp_of}
%   \begin{array}{rcl}
%   \displaystyle \min   & \hspace{-0.5em} ||\beta -\tilde{\beta}||^2 + \delta ||\bflambda||^2 \,,\\
%   \textrm{s.t.}        & \hspace{-3.8em} \bfF_1,\bfF_2 \succeq 0 \,.
%   \end{array}
% \end{equation}
\begin{equation}
  \label{eq:sdp_of}
  \begin{array}{rclll}
  \displaystyle \min   & \hspace{-0.5em} ||\beta -\tilde{\beta}||^2 + \delta ||\bflambda||^2 \,,
  & \textrm{s.t.}        & \bfF_1,\bfF_2 \succeq 0 \,.
  \end{array}
\end{equation}
Now, the error will converge with a speed imposed by $\tilde{\beta}$, if possible.
Otherwise, the constraint will be relaxed so the system is stable ($\bfF_1$) and the Euler approximation in \eqref{eq:taylor_error} holds ($\bfF_2$).
Setting a high value for $\tilde{\beta}$ might be seen as maximizing the speed convergence while keeping the stability and the joints' velocity limits.
Notice that, for the sake of the problem solvability, it is also necessary to add a regularization term $\delta$ related to the gains.

In order to convert~\eqref{eq:sdp_of} into an LMI we must provide a linear objective function.
This can be done by upper-bounding the quadratic expression with an additional optimization variable $\gamma \inR$.
Thus, we will minimize $\gamma$ subject to the following constraint
% \begin{equation}
%   \xcancel{\gamma - (\bflambda -\tilde{\bflambda})^\top(\bflambda -\tilde{\bflambda}) \geq 0 \,.}
% \end{equation}
% \vspace{-0.2em}
\begin{equation}
% \vspace{-0.2em}
  \label{eq:optim_gamma}
  \gamma - (\beta -\tilde{\beta})^2 - \delta \bflambda^\top\bflambda \geq 0 \,.
\end{equation}
Again, we can express the constraint~\eqref{eq:optim_gamma} as an LMI by taking advantage of the Schur complement
% \begin{equation}
%   \xcancel{\bfF_3(\bflambda, \gamma) =
%   \begin{bmatrix}
%     \gamma - 2\tilde{\bflambda}^\top\bflambda & \bflambda^\top \\
%     \bflambda & \bfI  \\
%   \end{bmatrix}
%   \succeq \bf0\,.}
% \end{equation}
% \vspace{-0.2em}
\begin{equation}\label{eq:F3}
% \vspace{-0.2em}
  \bfF_3(\bflambda, \beta, \gamma) =
  \begin{bmatrix}
    \gamma & \bflambda^\top & \beta - \tilde{\beta} \\
    \bflambda & \delta^{-1}\bfI & \bf0 \\
    \beta - \tilde{\beta} & \bf0 & 1
  \end{bmatrix}
  \succeq \bf0\,.
\end{equation}
Notice that an extra constraint to guarantee $\beta >0$ is also necessary.
For the sake of conciseness and due to its simplicity, it has not been detailed here.

By introducing $\beta$ and $\gamma$ in the problem, we are adding two new components to the vector of decision variables, hence 
% \sout{${\bfx = [\bflambda^\top, \gamma]^\top}$} 
${\bfx = [\bflambda^\top, \beta, \gamma]^\top}$.
Therefore, two extra matrices $\bfF_{3,n+1}$ and $\bfF_{3,n+2}$ should be added in the computation of $\bfF_3$ in \eqref{eq:single_LMI}, which will be multiplied by $\beta$ and $\gamma$, respectively. 
% \sout{, and where all of its elements are zero in exception of the element in the first column of the first row, which is $1$}.
Notice how the previous LMIs (${\bfF_{1},\bfF_{2}}$) do not depend on $\gamma$ nor $\beta$ and their corresponding $\bfF_{\{j,n+1\}}$ matrices can be omitted since they are null.
With $\gamma$ the variable to be minimized, we can define the coefficient vector $\bfc$ of the cost function in \eqref{eq:SDP_standard_form} as a vector with zeros in the positions related to each component of $\bflambda$ and $\beta$ (\ie $n+1$ elements with 0) and a one in the position of the upper bound $\gamma$, hence ${\bfc = [{\bf0}_{1\times n+1},1]^\top \inR^{n + 2}}$.

% \subsubsection{Final SDP}

% In summary, considering the described LMIs \ie ${\bfF_j \text{ with } j \in [1,2,3]}$, the SDP problem formulation shown in \eqref{eq:SDP_standard_form} results in
% % \begin{equation}
% % \label{eq:SDP_resulting_form}
% % \begin{array}{rl}
% % \displaystyle \min_{\bfx} 	&  \bfc^\top\bfx \\
% % \textrm{s.t.}             	& \mathsf{blockdiag}(\bfF_1(\bfx), \bfF_2(\bfx), \bfF_3(\bfx))  \succeq 0 \,.
% % \end{array}
% % \end{equation}
% \begin{equation}
% \label{eq:SDP_resulting_form}
% \begin{array}{rlll}
% \displaystyle \min_{\bfx} 	& \hspace{-0.6em} \bfc^\top\bfx & 
%  \hspace{-0.6em} \textrm{s.t.}             	&  \hspace{-0.6em} \mathsf{blockdiag}(\bfF_1(\bfx), \bfF_2(\bfx), \bfF_3(\bfx))  \succeq 0 \,.
% \end{array}
% \end{equation}
%%%%%%%%%%%%%%%%%%%%%%%%%%%%%%%%%%%%%%%%%%%%%%%%%%%%%%%%%%%%%%%%%%%%%%%%%%%%%%%%
%%%%%%%%%%%%%%%%%%%%%%%%%%%%%%%%%%%%%%%%%%%%%%%%%%%%%%%%%%%%%%%%%%%%%%%%%%%%%%%%
%%%%%%%%%%%%%%%%%%%%%%%%%%%%%%%%%%%%%%%%%%%%%%%%%%%%%%%%%%%%%%%%%%%%%%%%%%%%%%%%

\section{Validation}
\label{sec:validation}
Task stabilization through the optimization procedure presented in this work is of use when dealing with highly redundant robots that must perform several tasks.
In these cases, the analytical study to choose the right gains, hence to guarantee stability, becomes unfeasible.
% \sout{(\ie when the number of variables to consider is high)}
% \comb{In this kind of robots, the analytical analysis becomes too complex. 
When the number of tasks and required DOFs are not high (\eg two tasks and three DOFs), this method is also of use as it eases the gains search that will render all tasks stable, without the need for simplifications as in other existing methods (\eg it allows to consider different gains for each task dimension).

% \sout{However, u}

The effectiveness of the mathematical developments proposed in this work can be better explained with robots with a low number of DOFs.
% Hence, without loss of generality and similar to~\cite{antonelli_stability_2009}, we first present a numerical experiment with a 3-link planar manipulator performing two tasks.
% Then, we include the simulations using the commercial UR5\footnote{\scriptsize\url{https://www.universal-robots.com/products/ur5-robot/}} robot arm performing with the on-line task gain tuning approach.
Hence, without loss of generality, we present a numerical experiment with the commercial UR5 (6 DOFs) robotic arm performing with the on-line task gain tuning approach.
These examples suffice to validate
% \sout{the optimization solution adopted and the role of every LMI presented in the previous section.}
that:
\begin{enumerate}
	\item[a)] In all simulations the stability condition in~\eqref{eq:stability_condition} holds. 
	It can be checked by looking at the Lyapunov function, which has to be monotonically descreasing.
	\item[b)] Variations on $\tilde{\beta}$ truly affect the convergence speed. For higher values of $\tilde{\beta}$ the tasks should converge faster.
	\item[c)] Joint velocity bounds $\overline{\dbfq}, \underline{\dbfq}$ are respected and the system still manages to converge by relaxing $\beta$.
	\item[d)] The method can handle different values of $\Delta t$ without affecting the system stability.
\end{enumerate}

We have performed the simulations by taking advantage of \texttt{Matlab} and the toolbox presented in \cite{corke_robotics_2017} to simulate a robot manipulator.
Moreover, we use the already existing SDP solver \texttt{Sedumi}~\cite{sturm_using_1999}.
All code related with this paper is made publicly available for the benefit of the community%
\footnote{\scriptsize\url{https://gitlab.iri.upc.edu/jmarti/SDP_HierarchicalTaskStability}}.

We have set a use case with the aim of reproducing a hand-writing action done by a human arm.
Thus, we set a primary task to follow a desired position path with the robot's end effector, whose error is described by $\tilde{\bfsigma}_1\inR^3$.
Besides, we impose the wrist to be close to the writing surface as a secondary task, \ie we impose the $y$ coordinate of the 4th joint to have a specific value ($\tilde{\sigma}_2\inR$).

The parameters for this case study are \mbox{$\bfq_0 = [135, 0, -90, 0, 90, 0]^\top \text{deg}$},
\mbox{$\bfsigma_{1}^\ast = [-0.5,-0.4,0.6]^\top \text{m}$}
and \mbox{$\sigma_{2}^\ast  = -0.3 \text{m}$}.
The regularization parameter is $\delta=5\times 10^{-5}$.

We have performed several experiments to validate the items (a)-(d) stated above.
Although the system stability (a) has been confirmed in all simulations, for the sake of conciseness, it is only reported here the figure that shows the stability of the experiments where we want to show the effectiveness of $\tilde{\beta}$.
In those cases, we have set $\Delta t = 0.01\text{s}$ and $\overline{\dbfq} = -\underline{\dbfq} = \bf6 \text{ rad/s}$ (we have used a boldface number to indicate the same limit for all the robot's joints) and ran several simulations to compare our method with both $\tilde{\beta} = 2$ and $\tilde{\beta} = 8$ against the tasks deployment with $\bflambda = [2,2,2,1]^\top$ as fixed gains.
For the sake of readability, all plots for this robot type are time-cropped to $4\text{s}$.

\begin{figure}[t!]
    \vspace{-1em}
	\includegraphics[width=0.9\columnwidth, angle=0, trim={30 20 80 20},clip]{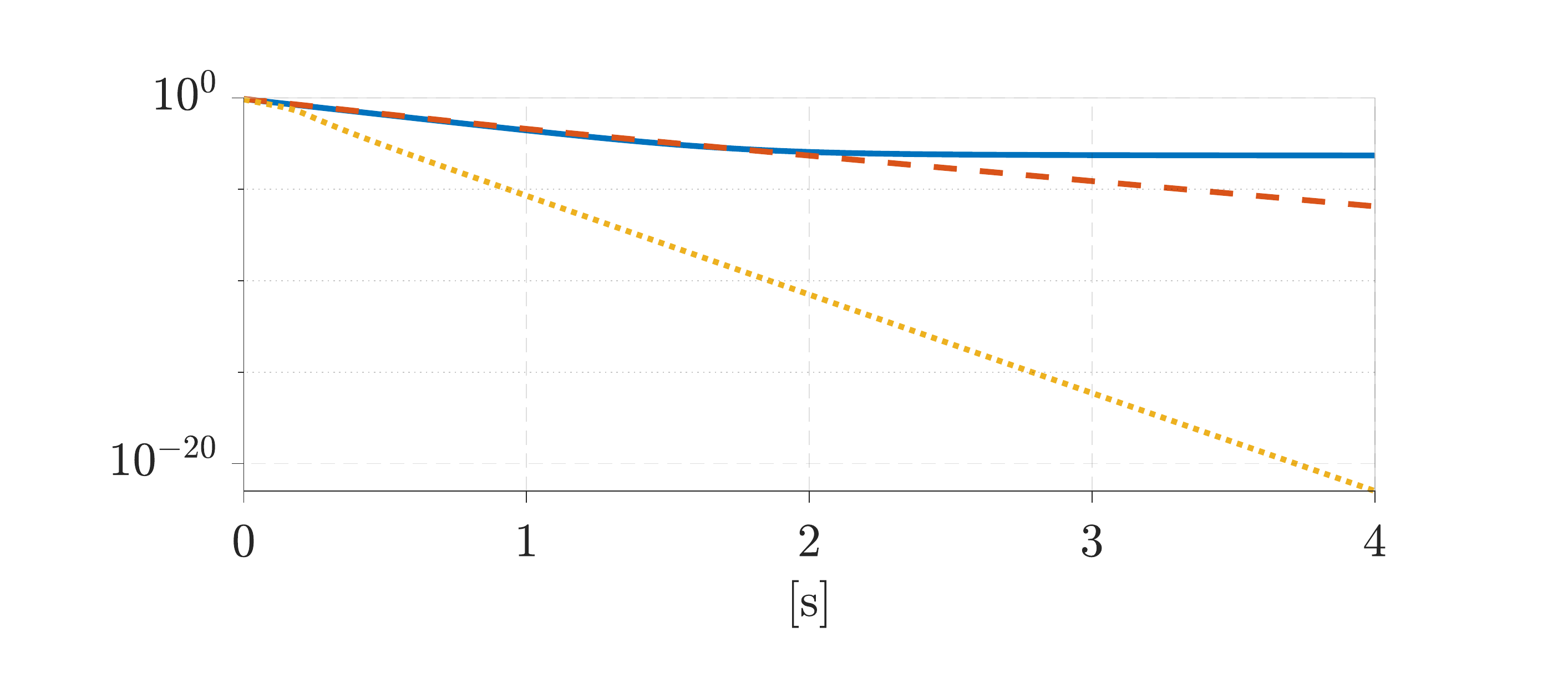}
	\vspace{-1em}
	\caption{Lyapunov function considering different values of $\tilde{\beta}$. Constant gains (solid blue), $\tilde{\beta} = 2$ (dashed red) and $\tilde{\beta} = 8$ (dotted yellow).}
	\label{fig:ur5_lyap}
	\vspace{-2em}
\end{figure}

\begin{figure}[t]
\vspace{-1em}
  \begin{tabular}{lr}
      \rotatebox{90}{\hspace{-2.5em}a) 1st task error.} & \hspace{-1.5em}
\begin{subfigure}{.5\textwidth}
	\includegraphics[width=0.8\columnwidth, angle=0, trim={30 20 80 20},clip]{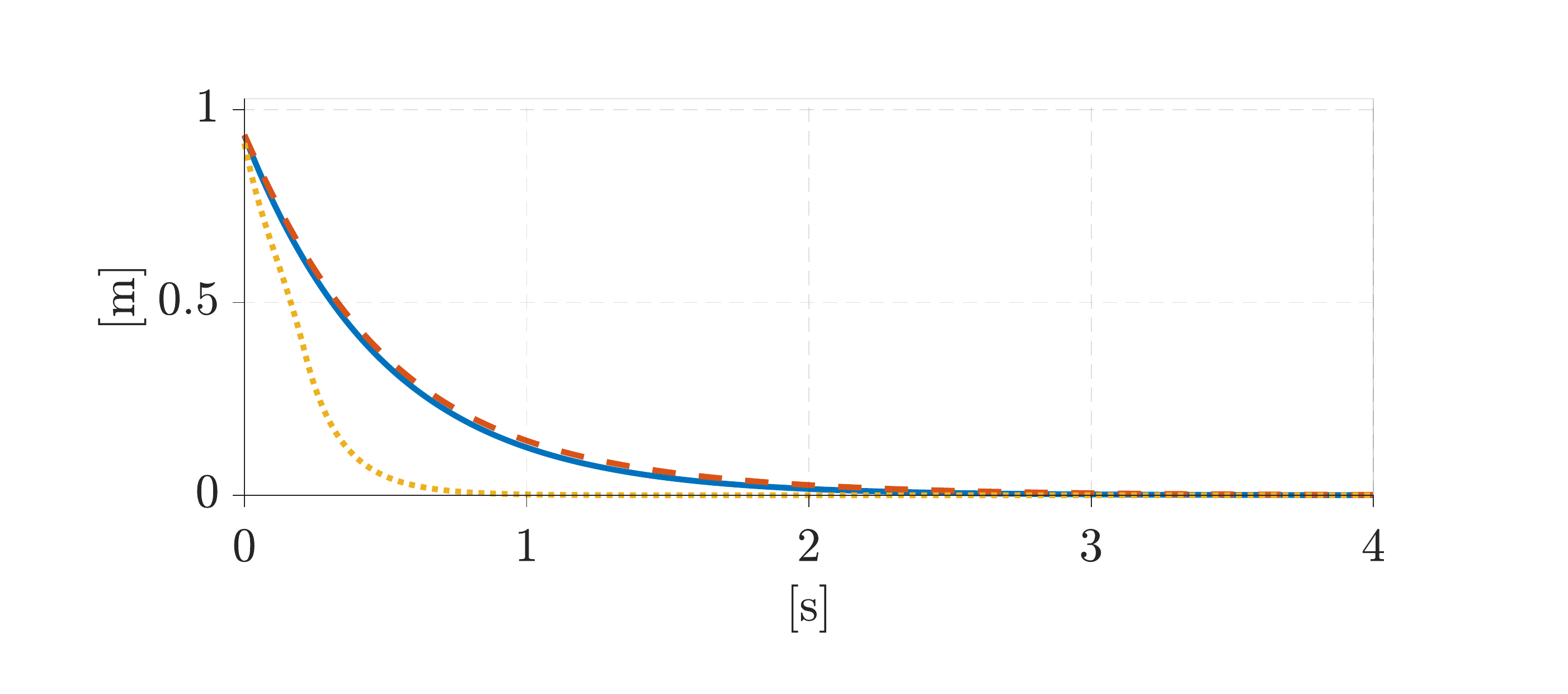}
	\captionlistentry{}
	\label{fig:ur5_eepos}
\end{subfigure} \vspace{-0.5em} \\ \vspace{-0.5em}
      \rotatebox{90}{\hspace{-3em}b) 2nd task error.} & \hspace{-1.5em}
\begin{subfigure}{.5\textwidth}
	\includegraphics[width=0.8\columnwidth, angle=0, trim={30 20 80 20},clip]{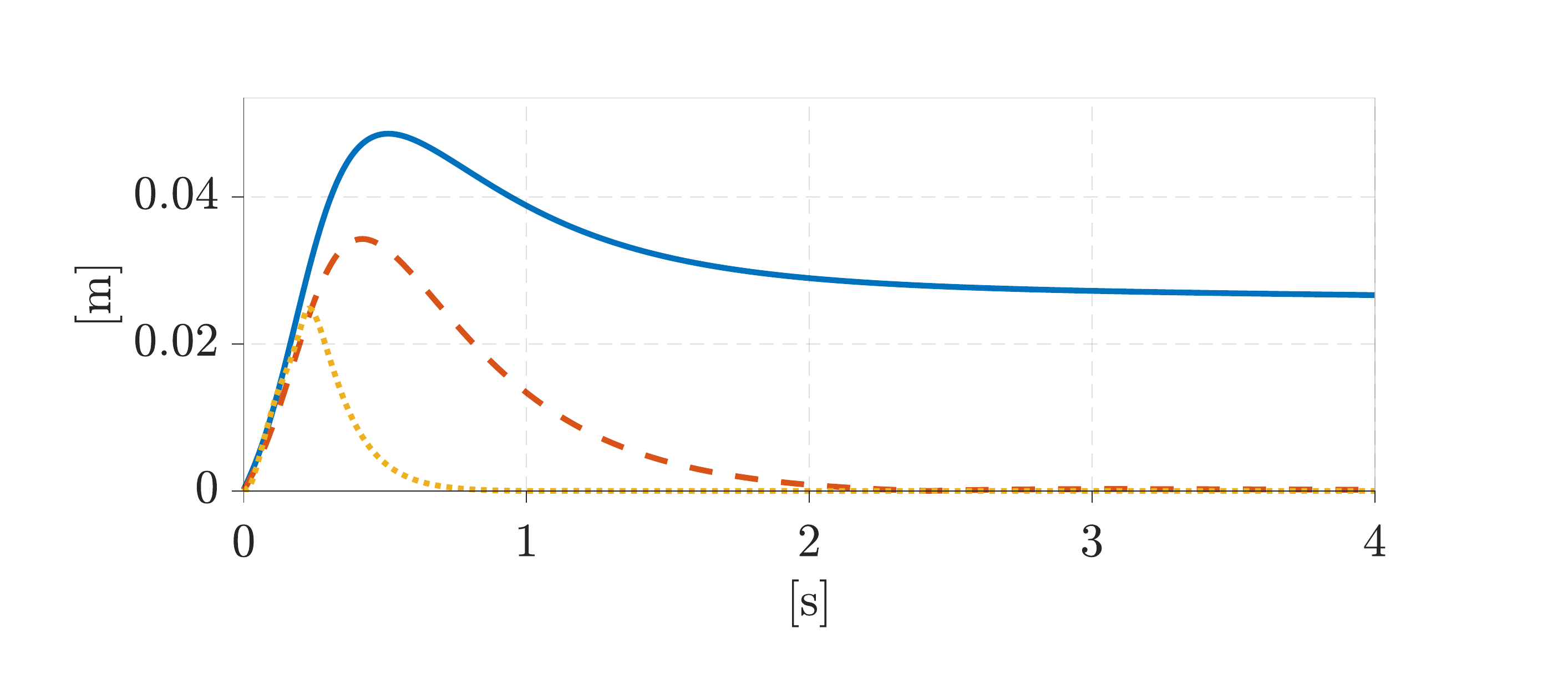}
	\captionlistentry{}
	\label{fig:ur5_eewr}
\end{subfigure}
\end{tabular}
\vspace{-0.5em}
\caption{Tasks error norms considering different values of $\tilde{\beta}$. Constant gains (solid blue), $\tilde{\beta} = 2$ (dashed red), $\tilde{\beta} = 8$ (dotted yellow).}
\label{fig:ur5_tasks}
\vspace{-0.5em}
\end{figure}

The corresponding results are shown in \figref{fig:ur5_lyap}, where the stability of the CLIK method is not guaranteed when using constant $\bflambda$ gains.
Notice that, while for the case of using the presented method the Lyapunov function decreases monotonously whereas for the case with constant gains the Lyapunov function remains stable once the first task has converged.
The direct consequence of not having a proper gain tuning method is that tasks with low priority are less prone to converge (see \figref{fig:ur5_eewr}).

In the example, we see that a high $\tilde{\beta}$ translates to faster error convergence for all tasks.
This effect can be seen in \figref{fig:ur5_lyap} and \figref{fig:ur5_tasks} for the primary and secondary tasks.


\begin{figure}[t!]
    \vspace{-1em}
	\includegraphics[width=0.9\columnwidth, angle=0, trim={30 20 80 20},clip]{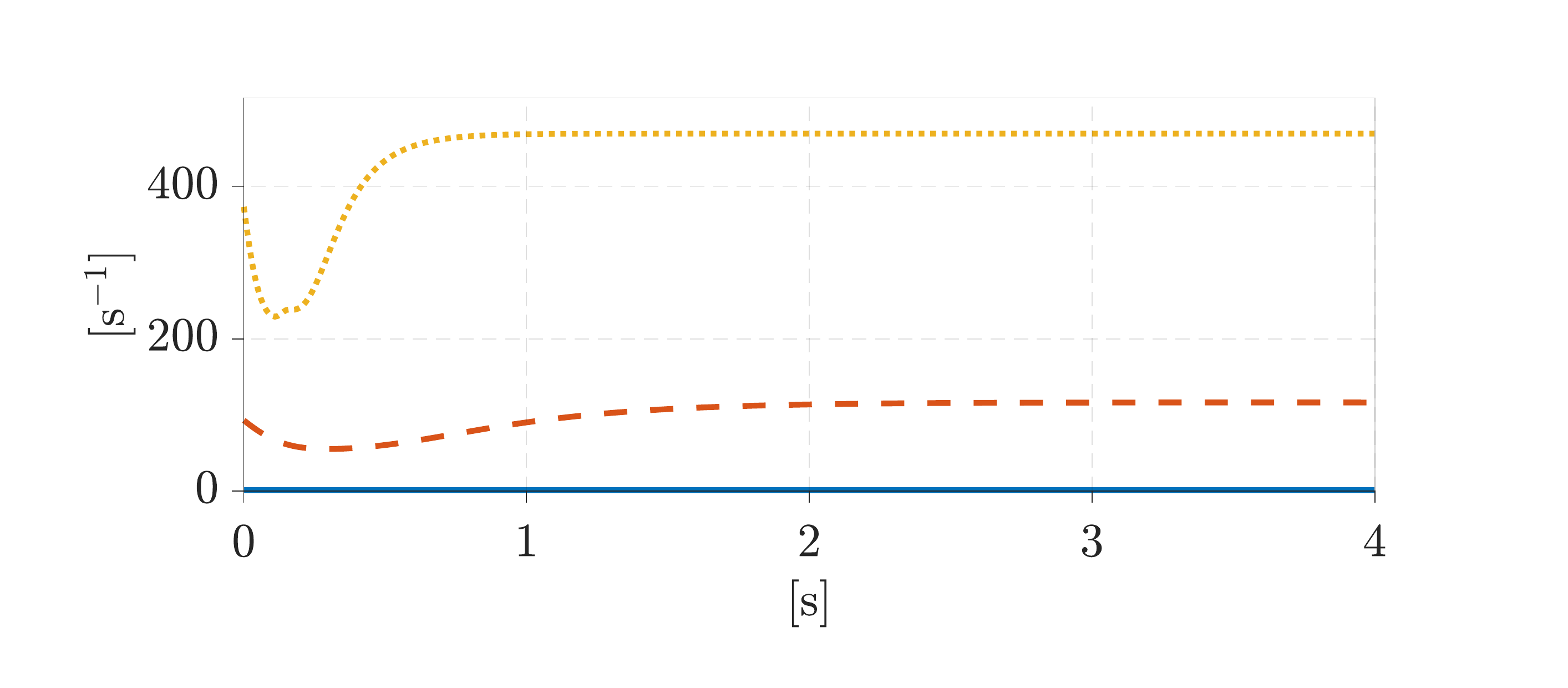}
	\vspace{-1em}
	\caption{Gain associated to the second task considering different values of $\tilde{\beta}$. Constant gains (solid blue), $\tilde{\beta} = 2$ (dashed red) and $\tilde{\beta} = 8$ (dotted yellow).}
	\label{fig:ur5_Kwr}
	\vspace{-2em}
\end{figure}

In this experiment, the manipulator is driven towards a configuration where the two tasks are close to be dependant, \ie where the rank of the augmented Jacobian is smaller than the sum of ranks of the respective Jacobians.
In such a situation, the null-space of the first task becomes smaller, hence the joint velocities devoted to fulfill the low priority task become smaller.
When operating in a close-loop manner, this vanishing joint velocity can be tackled by increasing the gains,
a benefit of imposing~\eqref{eq:stability_condition_mod_2} within our SDP approach.
This effect is shown in~\figref{fig:ur5_Kwr}, where the orientation task gain is increased as the configuration approaches the singularity.

\begin{figure}[t!]
  % \smallskip
    \vspace{-1em}
	\includegraphics[width=0.9\columnwidth, angle=0, trim={30 20 80 20},clip]{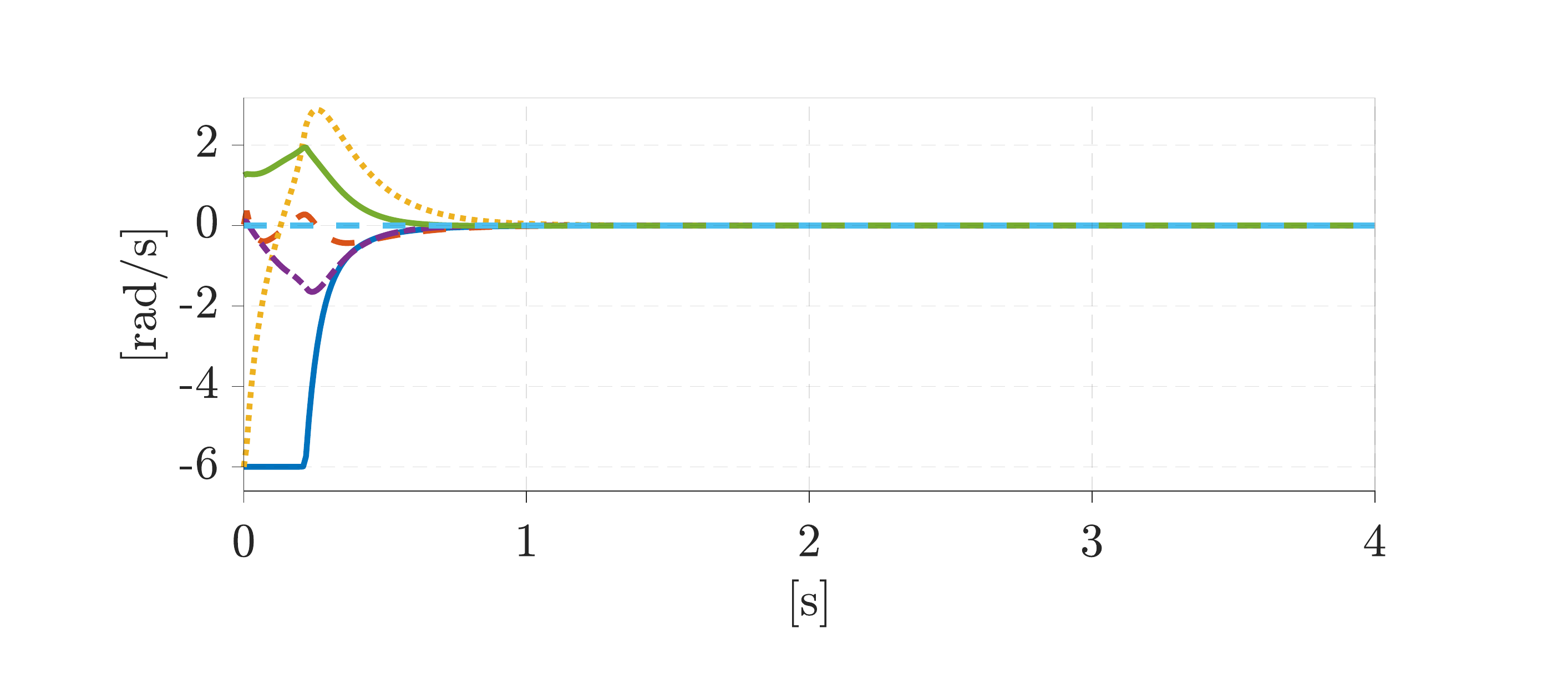}
	\vspace{-1em}
	\caption{Limited joint velocities when $\tilde{\beta}=8$ and $\overline{\dbfq}=-\underline{\dbfq} = \bf6 \text{ rad/s}$. Colors related to each joint in increasing order: solid blue, dashed red, dotted yellow, dashed magenta, solid green, dashed cyan.}
	\label{fig:ur5_jointvel}
	\vspace{-0.5em}
\end{figure}

\begin{figure}[t!]
    \vspace{-1em}
	\includegraphics[width=0.85\columnwidth, angle=0, trim={30 20 80 20},clip]{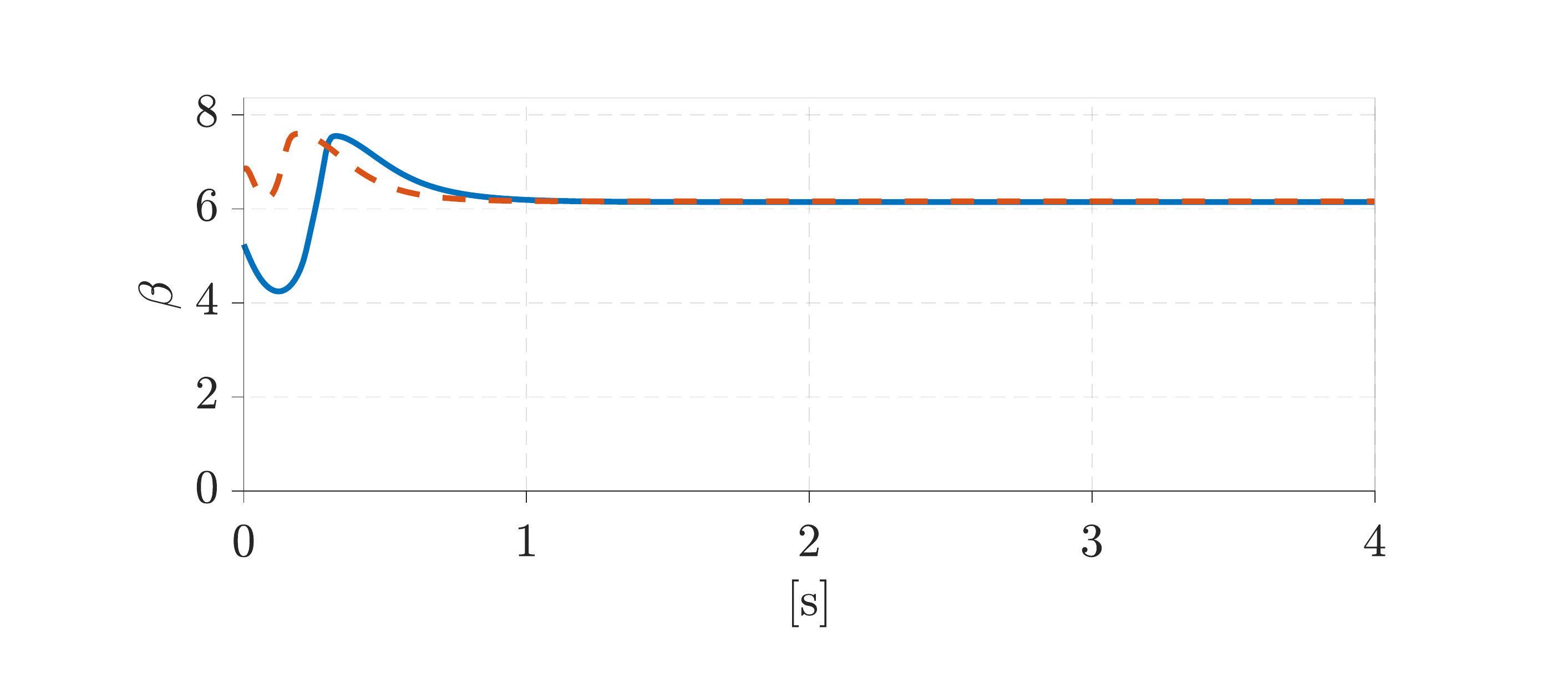}
	\vspace{-1em}
	\caption{$\beta$ value for different values of $\overline{\dbfq}$ and $\underline{\dbfq}$. Fixed value of $\tilde{\beta} = 8$.  $\overline{\dbfq}=-\underline{\dbfq}= \bf4 \text{ rad/s}$ (solid blue), $\overline{\dbfq}=-\underline{\dbfq} = \bf6 \text{ rad/s}$ (dashed red).}
	\label{fig:ur5_beta}
	\vspace{-2em}
\end{figure}

As previously noted, imposing faster error dynamics by increasing $\tilde{\beta}$ results in increasing the joint velocities.
This fact might violate the joint velocity constraint ($\bf F_2$) if $\beta$ was not properly relaxed ($\bf F_3$).
This case is shown in~\figref{fig:ur5_jointvel}, where the limits for the joints velocities are respected while the $\beta$ value is moved away from its desired $\tilde{\beta}$ (\figref{fig:ur5_beta}), in order to keep the joints velocity limits while guaranteeing the system stability.
\figref{fig:ur5_beta} shows the impact on $\beta$ while imposing different joint velocity limits, \ie the more restricting joint velocities (solid blue line), the further $\beta$ has to be moved from its desired value.

\begin{figure}[t!]
    % \vspace{-0.5em}
	\includegraphics[width=0.85\columnwidth, angle=0, trim={30 20 80 20},clip]{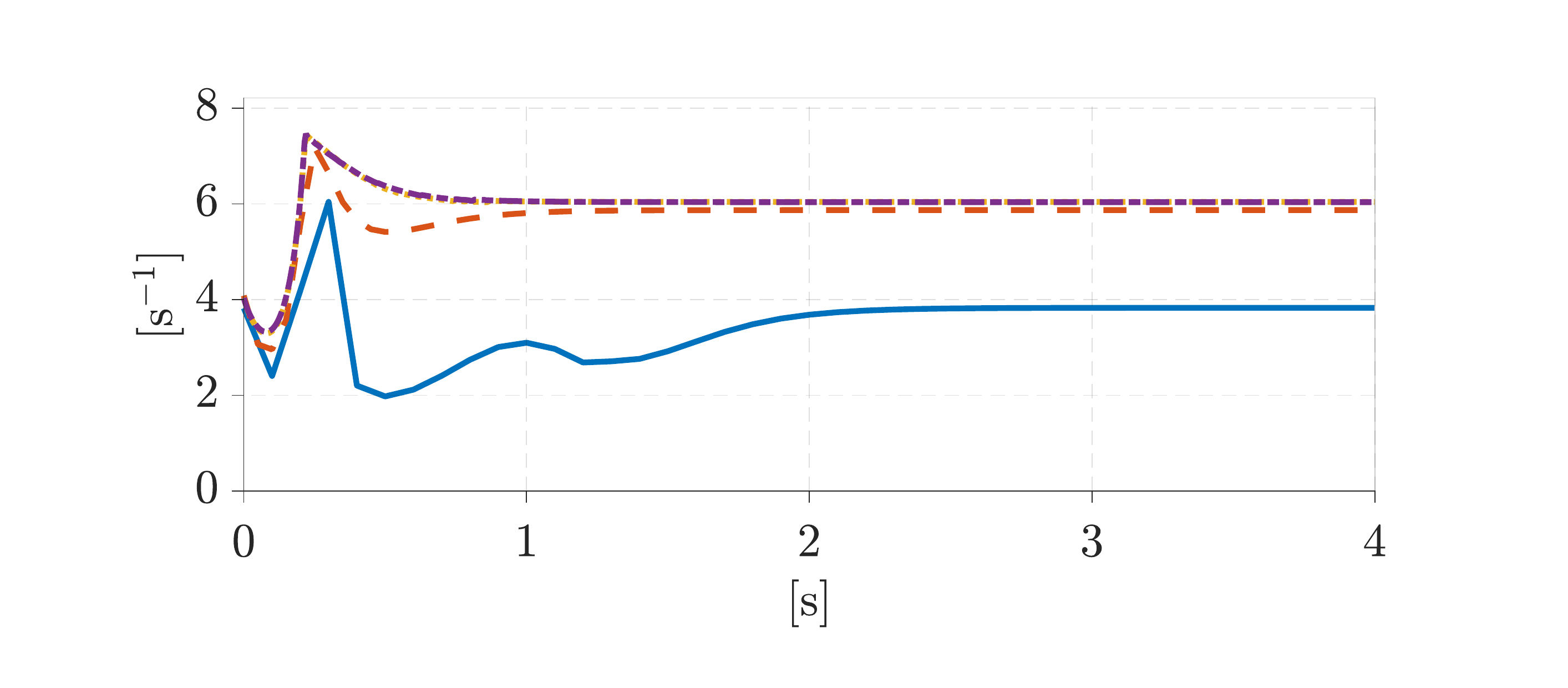}
	\vspace{-1em}
	\caption{$\lambda_1$, \ie gain associated to the first task's $x$-coordinate, for different $\Delta t$. $\Delta t = 0.1$s (solid blue), $\Delta t = 0.05$s (dashed red), $\Delta t = 0.01$s (dotted yellow), $\Delta t = 0.005$s (dashed magenta).}
	\label{fig:ur5_Kpos1_dt}
	\vspace{-1.0em}
\end{figure}

\begin{figure}[t!]
    \vspace{-1em}
	\includegraphics[width=0.85\columnwidth, angle=0, trim={30 20 80 20},clip]{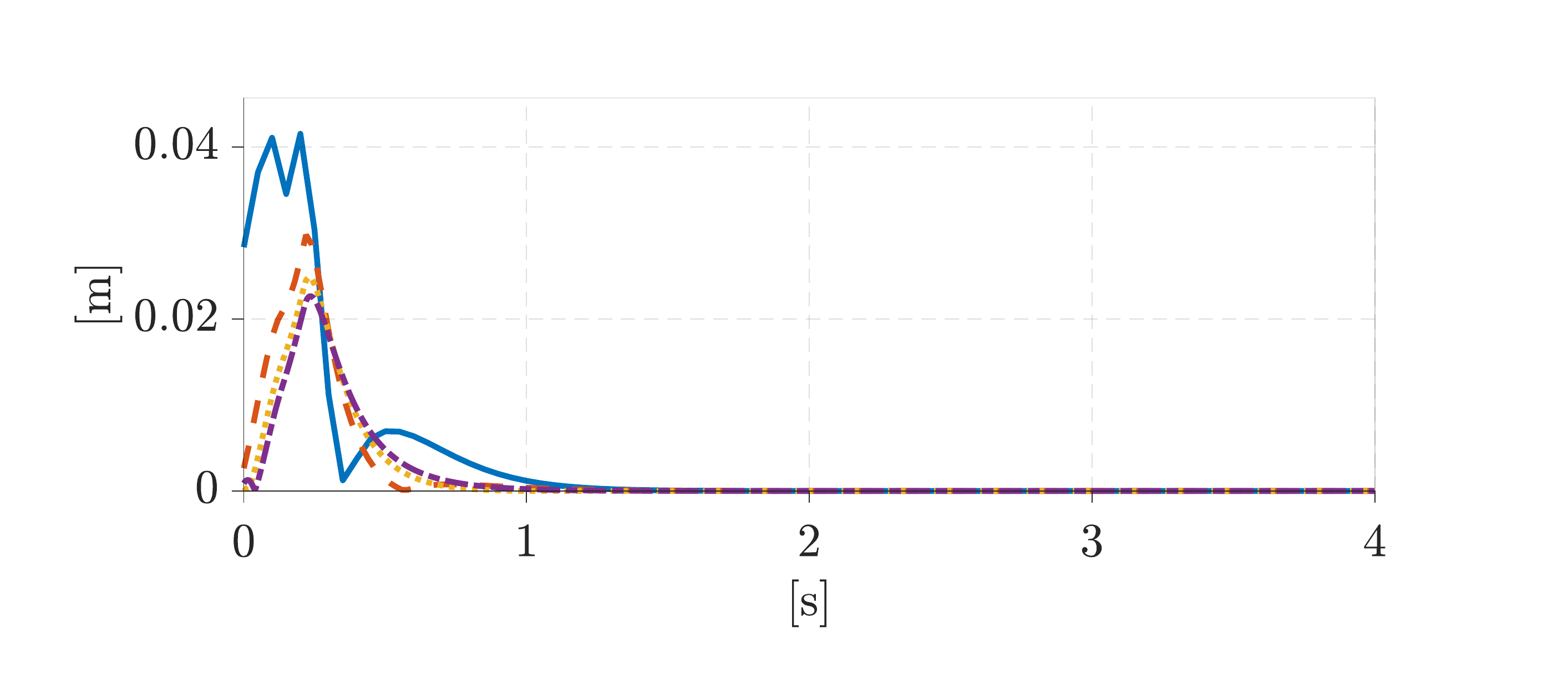}
	\vspace{-1em}
	\caption{Second task norm error for different $\Delta t$. $\Delta t = 0.1$s (solid blue), $\Delta t = 0.05$s (dashed red), $\Delta t = 0.01$s (dotted yellow), $\Delta t = 0.005$s (dashed magenta).}
	\label{fig:ur5_eewr_dt}
	\vspace{-2.5 em}
\end{figure}

The effect of the sampling time in the system performance is depicted in~\figref{fig:ur5_Kpos1_dt} and~\figref{fig:ur5_eewr_dt}.
We have set the desired speed to $\tilde{\beta}=8$ and the joint velocity limits to $\overline{\dbfq} = -\underline{\dbfq} = \bf6 \text{ rad/s}$ across all different $\Delta t$.
As with previous experiments, the method manages to stabilize the two tasks for all the tested $\Delta t$, \ie in all cases the maximum eigenvalue remains negative.
Therefore, as shown in~\figref{fig:ur5_Kpos1_dt}, the different task gains must be adapted (we could not notice a remarkable difference for the second task's gain).
Besides, the speed convergence of the second task improves with smaller sampling time (see~\figref{fig:ur5_eewr_dt}).
An important remark is that for a sufficiently small $\Delta t$, the quadratic term vanishes and~\eqref{eq:stability_condition_mod_2} becomes the stability condition of the continuous-time CLIK algorithm.
This fact is shown in both~\figref{fig:ur5_Kpos1_dt} and~\figref{fig:ur5_eewr_dt}, where the trajectories converge to a specific solution as $\Delta t$ decreases.


\section{Discussion and conclusions}
\label{sec:conclusions}
The focus of this work has been the mathematical development to introduce the use of an SDP approach with its advantages with respect to state-of-art methods.
However, some issues still remain open for further investigation.
First, since this is a problem of local stability, the error has an associated region of attraction which should be estimated.
In \cite{falco_stability_2011}, a method is proposed to estimate this region for a single task, however, an estimation for multiple tasks still remains as an open problem.
% , hence we limited the scope of the paper to guarantee local stability.
Second, the development of the stability condition is based upon an Euler approximation of the error \eqref{eq:taylor_error}, which is valid for a sufficiently small value of $||\dbfq||\Delta t$.
An upper bound for this product should be found.
Third, an open issue is how this method can be adapted to work in the acceleration domain so as to consider actuator torque limitations.
Finally, even though SDP techniques using LMIs offer the possibility of using fast and dedicated solvers, our method is based on iterative procedures (\ie optimization), which in contrast to analytical solutions requires some extra effort in efficient programming to run it in real-time.

The authors would like to thank Gianluca Antonelli for his advise in producing the final manuscript version.

% {\footnotesize
% \section{Acknowledgements}
% \label{sec:acknowledgements}
% The authors would like to thank Gianluca Antonelli for his advise in producing the final manuscript version.
% % }

%%%%%%%%%%%%%%%%%%%%%%%%%%%%%%%%%%%%%%%%%%%%%%%%%%%%%%%%%%%%%%%%%%%%%%%%%%%%%%%%
%%%%%%%%%%%%%%%%%%%%%%%%%%%%%%%%%%%%%%%%%%%%%%%%%%%%%%%%%%%%%%%%%%%%%%%%%%%%%%%%
%%%%%%%%%%%%%%%%%%%%%%%%%%%%%%%%%%%%%%%%%%%%%%%%%%%%%%%%%%%%%%%%%%%%%%%%%%%%%%%%

\addtolength{\textheight}{-12cm}   % This command serves to balance the column lengths

%%%%%%%%%%%%%%%%%%%%%%%%%%%%%%%%%%%%%%%%%%%%%%%%%%%%%%%%%%%%%%%%%%%%%%%%%%%%%%%%

\bibliographystyle{IEEEtran}
\balance
\bibliography{IEEEabrv,./files/references}

\end{document}